\documentclass{elsarticle}

\usepackage{lineno,hyperref}
\modulolinenumbers[5]

\usepackage{multirow}
\usepackage[flushleft]{threeparttable}
\usepackage[table]{xcolor}
\usepackage{changepage}

\usepackage{graphicx}
\graphicspath{ {./img/} }
\usepackage{color, colortbl}
\definecolor{Gray}{gray}{0.9}
\definecolor{blue}{rgb}{0.6356,0.7242,0.8775}

\usepackage{amsmath}
\usepackage{amssymb}
\usepackage{caption}
\usepackage{subcaption}

\newcommand{\R}{\mathbb{R}} 
\newcommand{\V}[1]{\mathbf{#1}} 

\DeclareMathOperator*{\argmax}{argmax}

\begin{document}

\begin{frontmatter}

\title{Scope resolution of predicted negation cues: A two-step neural network-based approach}

\author[1]{Daan de Jong \fnref{fn1}}
\ead{d.dejong1@uu.nl}

\address[1]{Utrecht University, Department of Methodology and Statistics, The Netherlands}

\begin{abstract}
Neural network-based methods are the state of the art in negation scope resolution. However, they often use the unrealistic assumption that cue information is completely accurate. Even if this assumption holds, there remains  a dependency on engineered features from state-of-the-art machine learning methods. The current study adopted a two-step negation resolving apporach to assess whether a Bidirectional Long Short-Term Memory-based method can be used for cue detection as well, and how inaccurate cue predictions would affect the scope resolution performance. Results suggest that this method is not suitable for negation detection. Scope resolution performance is most robust against inaccurate information for models with a recurrent layer only, compared to extensions with a Conditional Random Fields layer or a post-processing algorithm. We advocate for more research into the application of deep learning on negation detection and the effect of imperfect information on scope resolution.
\end{abstract}


\end{frontmatter}

\section{Introduction} \label{intro}

Negation is a complex grammatical phenomenon that has received considerable attention in the biomedical Natural Language Processing (BioNLP) domain. Negations play an important role in the semantic representation of biomedical text, because they reverse the truth value of propositions \cite{SEM-shared-task}. Therefore, correct negation handling is a crucial step whenever the goal is to derive factual knowledge from biomedical text. 

We can distinguish two ways to approach negations in medical text: negation detection and negation resolving. Negation detection is a form of assertion identification, in this case, determining whether a certain statement is true or false, or whether a medical condition is absent or present \cite{mutalik2001, chapman2001, sanchez2007, huang2007, peng2017, bhatia2019, chen2019, sykes2020}. Negation resolving shifts the focus towards the token level by approaching the problem as a sequence labeling task \cite{morante2008}. This task is typically divided into two sub tasks: (1) detecting the negation \textit{cue}, a word expressing negation and (2) resolving its \textit{scope}, the elements of the text affected by it. A cue can also be a morpheme (``\textit{im}possible'') or a group of words (``not at all''). As an example, in the following sentence the cue is underlined and its scope is enclosed by square brackets: \begin{center}
``I am sure that [\underline{neither} apples \underline{nor} bananas are blue].'' 
\end{center}

Recently, researchers adopted a neural network-based approach to resolve negations. This approach is shown to be highly promising, but most methods solely focus on scope resolution, relying on gold cue annotations. As Read et al. \cite{read2012} point out: ``It is difficult to compare system performance on sub tasks, as each component will be affected by the performance of the previous.'' This comparison will not be easier when the performance on a sub task is not affected by the performance of the previous component.

The main advantage of deep learning methods is their independence of manually created features, in contrast to machine learning methods. However, by aiming at scope resolution only, they indirectly still use these features, or assume 100\% accurate cues. For complete automatic negation resolving, a neural network model should detect the cue by itself. This raises two questions:

\begin{enumerate}[(1)]
\item How would a neural network-based model perform on the cue detection task?
\item How would a neural network-based model perform on the scope resolution task with imperfect cue information?
\end{enumerate}

The current study addresses these questions by applying a Bi-directional Long Short-Term Memory (BiLSTM) model \cite{fancellu2016} to both stages of the negation resolving task. We develop the models on the BioScope Abstracts and Full Papers sub corpora \cite{bioscope}. The results suggest that word embeddings alone can detect cues reasonably well, but there still exist better alternatives for this task. As expected, scope resolution performance suffers from imperfect cue information, but remains acceptable on the Abstracts sub corpus. \newline 

\noindent As a secondary aim, the current study explores different methods to ensure continuous scope predictions. Since the BioScope corpus only contains continuous scopes, the Percentage Correct Scopes will likely increase after applying such a method. We compare a post-processing algorithm \cite{morante2008} with a Conditional Random Field (CRF) layer \cite{fancellu2017}. The results suggest that both methods are effective, although the post-processing negatively affects the token-based performance.

\section{Related Work}

\begin{table}[h!]
\centering
\small
	\begin{threeparttable}
	\caption{Performance of existing methods on two corpora.} \label{relatedworkatable}
	\begin{tabular}{| c | l | c | c | c |}
	\hline
	\multicolumn{5}{|c|}{\cellcolor{gray!50} Conan Doyle corpus \cite{conandoyle}}  \\ \hline
	
	\cellcolor{gray!25} Approach & \cellcolor{gray!25} Method & \cellcolor{gray!25} Cue det. F1 & \cellcolor{gray!25} Scope res. F1 & \cellcolor{gray!25} Cue input \\ \hline
	
	\multirow{2}{*}{RB} & Lexicon \cite{carrillo2012} & 90.26 & 76.03 & Pred \\ \cline{2-5}
	& Lexicon \cite{ballesteros2012} & 71.88 & 62.65 & Pred \\ \hline
	
	\multirow{4}{*}{ML} & Lexicon+SVM \cite{gyawali2012} & 85.77 & 76.23 & Pred \\ \cline{2-5}
	& SVM \cite{read2012} & 92.10 & 85.26 & Pred \\ \cline{2-5}
	& \multirow{2}{*}{MRS Crawler \cite{packard2014}}  & \multirow{2}{*}{-} & 86.6 & Gold \\ \cline{4-5}
	& & & 82.4 & Pred* \\ \hline
	
	\multirow{2}{*}{CRF} & CRF \cite{jbara2012} & 90.98 & 82.70 & Pred \\ \cline{2-5}
	& CRF \cite{white2012} & 90.00 & 83.51 & Pred \\ \hline
	
	\multirow{2}{*}{NN} & BiLSTM \cite{fancellu2016} & - & 88.72 & Gold \\ \cline{2-5}
	& NegBERT \cite{negbert} & \textbf{92.94} & \textbf{92.36} & Gold \\ \hline

	\multicolumn{5}{|c|}{} \\ \hline
	\multicolumn{5}{|c|}{\cellcolor{gray!50} BioScope Abstracts corpus \cite{bioscope}} \\ \hline
	
	\cellcolor{gray!25} Approach & \cellcolor{gray!25} Method & \cellcolor{gray!25} Cue det. F1 & \cellcolor{gray!25} Scope res. F1 & \cellcolor{gray!25} Cue input \\ \hline
	
	\multirow{4}{*}{ML} & \multirow{2}{*}{Memory-based \cite{morante2008}} & \multirow{2}{*}{91.54} & 88.40 & Gold \\ \cline{4-5}
	&	&	&	80.99 & Pred \\ \cline{2-5}
	& \multirow{2}{*}{Metalearner \cite{morante2009}} &  & 90.67 & Gold \\ \cline{4-5}
	&	&	\multirow{-2}{*}{\textbf{99.37}}  & 82.60 & Pred \\ \hline
	
	\multirow{4}{*}{NN} & CNN \cite{qian2016} & - & 89.91 & Gold \\ \cline{2-5}
	& BiLSTM+CRF \cite{fancellu2017} & - & 92.11 & Gold \\ \cline{2-5}
	& BiLSTM \cite{taylor2018} & NR & 88.85 & None \\ \cline{2-5}
	& NegBERT \cite{negbert} & 95.65 & \textbf{95.68} & Gold \\ \hline
	
	\end{tabular}
	\begin{tablenotes}
		\scriptsize
		\item Note: RB = Rule-based, ML = Machine Learning, CRF = Conditional Random Field, NN = Neural Networks. NR = Not Reported, a dash indicates that no cue detection was performed. *Predictions from SVM \cite{read2012}. 
	\end{tablenotes}
	\end{threeparttable}
\end{table}

Negation resolving has been tackled by a range of approaches: rule-based methods, Machine Learning (ML) and Conditional Random Fields (CRFs). In this section, we will briefly discuss these approaches, followed by a discussion of neural network-based studies. An brief overview of the performance of earlier proposed methods is provided in Table \ref{relatedworkatable}, see \ref{relatedworkappendix} for an extensive overview.  \newline

\noindent Rule-based methods were the first methods used for negation detection, but only later they were applied to negation resolving. Examples of rule-based approaches are the use of regular expression algorithms \cite{chapman2001, deepen}, pre-defined lexicons and syntax trees, \cite{carrillo2012, ballesteros2012} and text representations with formal semantic structures \cite{basile2012}. Within this approach, it is common to first detect the negation cues, and subsequently resolve their scope. 

Although rule-based methods show acceptable performance on both tasks, they do not easily generalize to other domains or even data sets. Machine Learning (ML) classifiers were introduced to overcome this problem, performing on par with or better than rule-based methods \cite{lapponi2012, cruz2015}. Examples are memory-based learning algorithms \cite{morante2008}, Support Vector Machines (SVM) \cite{gyawali2012}, metalearning approaches \cite{morante2009} and hybrid methods, combining SVM classifiers with heuristic rules \cite{read2012, packard2014}. Most ML methods are also designed for a two-step procedure where scope resulution is influenced by the accuracy of the cue predictions. Morante et al. \cite{morante2009} showed the importance of this problem by comparing their system with perfect and imperfect cue information, and reported a 8\% decrease in token-based F1 measure. Packard et al. \cite{packard2014} made a similar comparison and reported a 4\% F1 decrease when moving from gold cue annotations to predicted cue labels.

The two-step procedure was also adopted by researchers using Conditional Random Fields (CRF) models. These models are well suited for sequence labeling tasks, since a token sequence can be easily represented as a linear graph. Most of these models achieve acceptable performance on the scope resolution task with the use of predicted cue features and other syntactic features \cite{agarwal2010, jbara2012, white2012, li2018}.

Recently, researchers started to investigate the application of neural network models to scope resolution. In this way, hand-crafted features needed for Machine Learning could be replaced by unsupervised features. For example, Qian et al. \cite{qian2016} used Convolutional Neural Networks (CNNs) to extract path features and combined these with position features. BiLSTM-based models became the state of the art \cite{fancellu2016, fancellu2017, lazib2016}, capable of integrating word and cue embeddings into their memory cells. Later, Fei et al. \cite{fei2020} outperformed this method with a Recursive Neural Network that automatically learns syntactic features, combined with a CRF layer. All these methods aim at the scope resolution task, assuming gold cue information.

More recently, transformer-based models have shown to be the current state of the art \cite{negbert, britto2020}. Importantly, these models are also capaable of detecting negation cues. In the second stage of the task, they use a method that replaces the original token in the sentence by a special cue token. Currently, this stage is only performed with gold cue tokens. 

The tasks can also be solved separately, that is, by not passing information of the first sub task to the second. Gautam et al. \cite{gautam2018} developed an Encoder-Decoder LSTM for this approach. They showed that this model can detect negation cues with a 100\% precision in conversation data, using only word embeddings, and achieved near equal performance with simple one-hot word vectors. However, the model performed considerably worse on the scope resolution task. 

Serveega et al. \cite{sergeeva2019} recognized the dependency of neural network-based models on gold cue information, and proposed a BiLSTM-based model that achieved acceptable performance without using cue inputs. However, they do use Part-Of-Speech (POS) tags and dependency tree features. They compared model performance with gold cues, predicted cues and no cues and concluded that gold cues lead to the best performance, with little difference between predicted cues and no cues. For the cue predictions, they used an hierarchical LSTM model. Another method that did not use cue inputs was proposed by Taylor and Harabagiu \cite{taylor2018}. They tackled both tasks simultaneuously with a cue/outside/inside labeling scheme and showed that the BiLSTM still correctly identified 89.02\% of the scope tokens.

\section{Task modeling} 
Let a sentence be represented by a token sequence $\V{t}=(t_1~t_2~\cdots~t_n)$. Following Khandelwal and Sawant \cite{negbert}, we use the following labeling scheme for the cue detection task: For $k=1,\dots,n$, token $t_k$ token is labeled 

\begin{itemize}
\item \textbf{C} if it is annotated as a single word cue or a discontinuous multiword cue,
\item \textbf{MC} if it is part of a continuous multiword cue and
\item \textbf{NC} if it is not annotated as a cue.
\end{itemize}

\noindent The scope label of token $t_k$ token is

\begin{itemize}
\item \textbf{O} if it is outside of the cue's negation scope,
\item \textbf{B} if it is inside the negation scope, \textit{before} the first cue token, 
\item \textbf{C} if it is the first cue token in the scope and
\item \textbf{A} if it is inside the negation scope, \textit{after} the first cue token.\footnotemark
\end{itemize}

For each sentence, Task 1 is to predict its cue sequence $\V{c} = \{\mathbf{NC},\mathbf{C}, \mathbf{MC}\}^n$ given its token sequence $\V{t}$ and Task 2 is to subsequently predict the scope sequence $\V{s} = \{\mathbf{O}, \mathbf{B}, \mathbf{C},\mathbf{A}\}^n$ given $\V{t}$ and $\V{c}$. As an example, the token sequence $\V{t}$ with gold cue and scope labels of ``It had [\textbf{no} effect on IL-10 secretion].'' are given in Table \ref{example}.

\begin{table}[h!]
\centering
	\caption{Example of a token sequence and its cue and scope labels.}
	\label{example}
	\begin{tabular}{|l | c | c| c| c| c| c| c| c|}
	\hline
	\textbf{Tokens} & it 	& had 	& no 	& effect 	& on 	& IL-10 	& secretion 	& . \\
	\hline
	\textbf{Cue labels} & NC 		& NC 		& C 		& NC			& NC 		& NC 		& NC 			& NC \\
	\hline
	\textbf{Scope labels} & O 		& O 		& C 		& A 			& A 		& A 		& A 			& O \\
	\hline
	\end{tabular}
\end{table}

\footnotetext{See \ref{motivation} for a motivation of the scope labeling scheme.}

\subsection{Performance measures} 
To measure perfomance, we evaluate whether the tokens are correctly predicted as cue or noncue (Task 1) and as outside or inside the scope (Task 2). At the token level, both tasks are evaluated by precision, recall and F1 measures. 

At the scope level, we report the percentage of exact cue matches (PECM) over the number of negation sentences for Task 1. All cue tokens in the sentences have to be correctly labeled to count as an exact match. For Task 2, we adopt the Percentage of Correct Scopes (PCS) as a  measure of performance, the percentage of gold negation scopes that are completely match. To evaluate the effectiveness of a `smoothing' method, we compute the Percentage of Continuous Predictions (PCP) over all scope predictions.\footnote{Let the left and right boundary of a scope be defined as $k_L=\min\big\{k | s_k \in \{\mathbf{B}, \mathbf{C}, \mathbf{A}\}\big\}$ and $k_R=\max\big\{k | s_k \in \{\mathbf{B}, \mathbf{C}, \mathbf{A}\}\big\}$, respectively. We define a scope to be continuous if $t_k=1$ for all $k_L \le k \le k_R$, and discontinuous otherwise.}

\section{Model architecture}
\begin{figure}
\begin{adjustwidth}{-3cm}{}
	\includegraphics[scale=0.55]{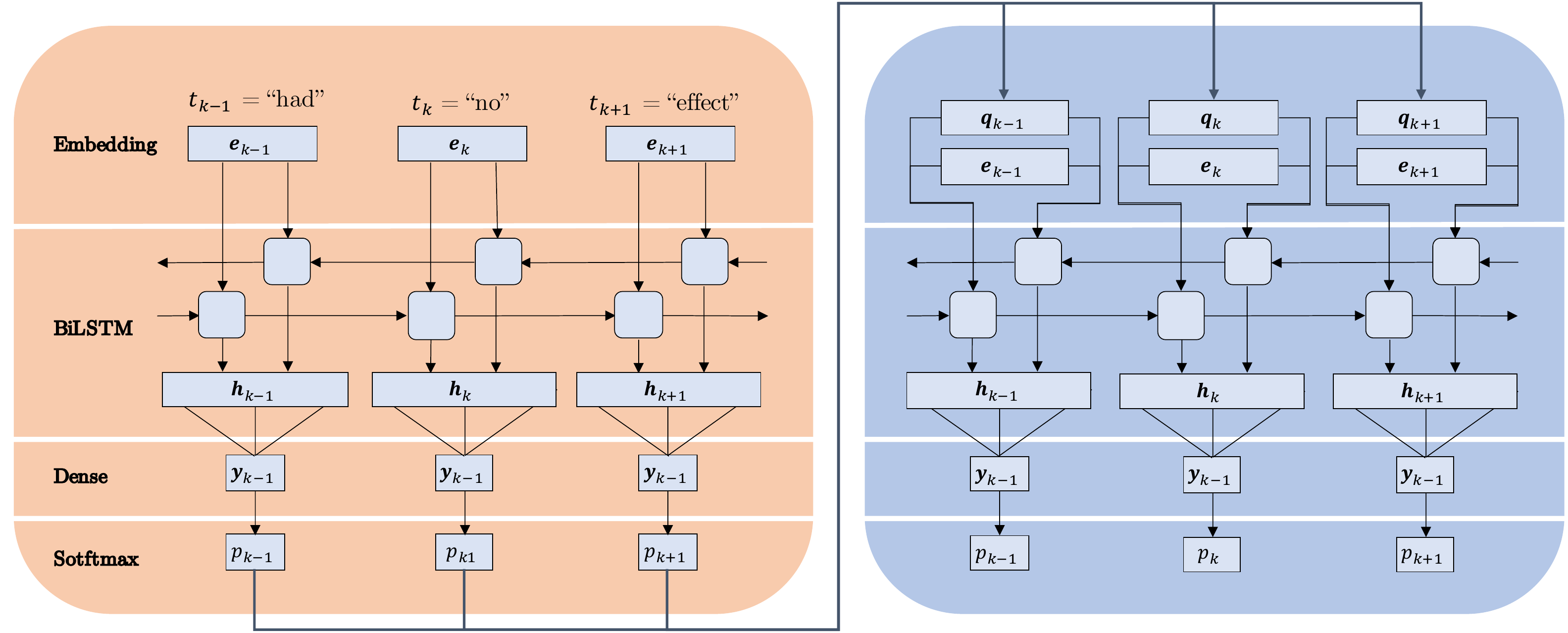}
	\caption{Schematic representation of the BiLSTM model for cue detection (left) and scope resolution (right), for the example sentence ``It had no effect on IL-10 secretion.'' at $k=3$.}
	\label{modelfigure}
\end{adjustwidth}
\end{figure}

In this section, we describe the proposed model architectures for Task 1 and Task 2. Both tasks are performed by a neural network consisting of an embedding layer, a BiLSTM layer and a softmax layer (Figure \ref{modelfigure}). For Task 1, we define a baseline model with an embedding layer and a softmax. For both tasks, we add a model where the softmax layer is replaced by a CRF layer to obtain a joint prediction for the token sequence. Finally, we discuss how the models were trained. 

\subsection{Word Embeddings for cue detection} 
The token sequence $\V{t}=(t_1~\cdots~t_n)$ is the only input for the cue detection models. Let  $E^{d\times v}$ be an embedding matrix, where $d$ is the embedding dimension and $v$ is the vocabulary size. Then, each token in $\V{t}=(t_1~\cdots~t_n)$ is represented by a pre-trained BioWordVec \cite{biosentvec} embedding $\V{e} \in \R^{d}$ corresponding to its vocabulary index. These embeddings were trained by the Fasttext subword embedding model with a context window size of 20 \cite{fasttext} on the MIMIC-III corpus \cite{mimicdata}. This model is able to include domain-specific subword information into its vector representations. Out-of-vocabulary (OOV) tokens were represented by a $d$-dimensional zero vector.

Word embeddings may represent features that are already informative enough for the cue detection task. Therefore, we define a baseline model where the embeddings are directly passed to a 3-unit dense layer with weights $W_s^{3\times d}$ and bias $\V{b}_s \in \R^3$. The output vector $$\V{y}_k =  W_{s} \V{e}_{k} + \V{b}_s = (y_k^{NC}, y_k^{C}, y_k^{MC})$$ contains to the `confidence' scores of tagging token $k$ as a noncue, cue or multiword cue, respectively. These scores are used to obtain the final prediction label $p_k = \mathrm{softmax}(\V{y}_k)$, where the softmax function $\R^3 \rightarrow \{\mathbf{NC}, \mathbf{C}, \mathbf{MC}\}$ is given by
$$\V{y} \mapsto \argmax\bigg\{\frac{e^{y^{NC}}}{Z}, \frac{e^{y^{C}}}{Z}, \frac{e^{y^{MC}}}{Z}\bigg\}, \quad Z=\sum_{y \in \V{y}} e^{y}.$$

\subsection{BiLSTM for cue detection} 
In the BiLSTM model, the token embeddings $(\V{e}_1~\cdots~\V{e}_n)$ are passed to a BiLSTM layer \cite{bilstm} with $2U$ units, $U$ in the forward direction and $U$ in the backward direction. We represent an LSTM layer as a sequence of $n$ identical cells. A cell at token $k$ is described by the following set of equations corresponding to its input gate $\V{i}_k$, forget gate $\V{f}_k$, output gate $\V{o}_k$, candidate memory state $\tilde{\boldsymbol{\gamma}}_k$, memory state $\boldsymbol{\gamma}_k$ and hidden state $\V{h}_k$, respectively:

$$
\begin{aligned}
\mathbf{i}_k &= \sigma\big(W_e^{(i)}\V{e}_k + W_h^{(i)}\V{h}_{k-1} +\V{b}^{(i)} \big), \\ 
\V{f}_k &= \sigma\big(W_e^{(f)}\V{e}_k + W_h^{(f)}\V{h}_{k-1} +\V{b}^{(f)} \big), \\
\V{o}_k &= \sigma\big(W_e^{(o)}\V{e}_k + W_h^{(o)}\V{h}_{k-1} +\V{b}^{(o)} \big), \\
\tilde{\boldsymbol{\gamma}}_k&=\mathrm{tanh}\big(W_e^{(\tilde{\boldsymbol{\gamma}})}\V{e}_k + W_h^{(\tilde{\boldsymbol{\gamma}})}\V{h}_{k-1} + \V{b}^{(\tilde{\boldsymbol{\gamma}})}\big), \\
\boldsymbol{\gamma}_k&=\mathbf{f}_k \odot \boldsymbol{\gamma}_{k-1} + \V{i}_k \odot \tilde{\boldsymbol{\gamma}}_{k}, \\
\V{h}_k&=\mathbf{o}_k \odot \mathrm{tanh}(\boldsymbol{\gamma}_k),
\end{aligned}
$$

where $W_e^{U \times d}$ denote the weight matrices for the token embeddings, $W_{h}^{U\times U}$ denotes the recurrent weight matrix, $\V{b} \in \R^{u}$ is a bias vector, $\odot$ denotes the Hadamard product, $\sigma$ denotes the sigmoid function\footnote{The function $\R \rightarrow (0,1)$ given by $x \mapsto 1/(1+e^{-x})$} and tanh denotes the hyperbolic tangent function \cite{hochreiter}.\footnote{The function $\R \rightarrow (-1,1)$ given by $x \mapsto (e^x-e^{-x})/(e^x+e^{-x})$} The hidden state of the forward layer and backward layer are concatenated to yield a representation $\overleftrightarrow{\V{h}}_{k}=(\overrightarrow{\V{h}}_k;\overleftarrow{\V{h}}_k) \in \R^{2u}$ for token $k$. For each token, the output $\overleftrightarrow{\V{h}}_{k}$ of the BiLSTM layer is fed into a 3-unit softmax layer with weights $W_s^{3\times 2U}$ and bias $\V{b}_s \in \R^3$, as defined in the baseline model.

\subsection{Adding CRF for cue detection}
Although the context around token $t$ is captured by the LSTM cell, the model will still assume independence between the token predictions when it maximizes a likelihood function. Alternatively, we can replace the softmax layer of the cue detection models by a Conditional Random Field (CRF) layer \cite{crf} to create a dependency between the predictions of adjacent tokens. This allows the model to learn that a single cue token is surrounded by noncue tokens, and that a multiword cue token is always followed by a next one. 

Let $Y = (\V{y}_1~\cdots\V{y}_n)$ be the $3\times n$ matrix of model predicted scores 
$$\begin{pmatrix} y_1^{NC} & y_2^{NC} & \cdots & y_n^{NC} \\ y_1^{C} & y_2^{C} & \cdots & y_n^{C} \\ y_1^{MC} & y_2^{MC} & \cdots & y_n^{MC} \end{pmatrix}.$$ Consider all possible label sequences enclosed by start/end labels $\mathcal{P} = \{\mathrm{start}\} \times \{\mathbf{NC},\mathbf{C},\mathbf{MC}\}^n \times \{\mathrm{end}\}$. Let $\V{p}^* \in \mathcal{P}$ and let $T \in \R^{5\times 5}$ be a matrix of transition scores, such that score $T_{i,j}$ corresponds to moving from the $i$-th to the $j$-th label in the set $\{\mathbf{NC},\mathbf{C},\mathbf{MC}, \mathrm{start}, \mathrm{end}\}$. Then, a linear CRF yields a joint prediction for a token sequence $\V{t}$ by attaching it a global score 
$$S(\V{t}, \V{c}, \V{p}^*) = \sum_{k=1}^{n} Y_{p_k^*, k} + \sum_{k=0}^{n} T_{p_k^*, p_{k+1}^*}.$$
The model predicts the label sequence with the maximum score among all possible label sequences:
$$\V{p} = \argmax_{\V{p}^*\in \mathcal{P}} S(\V{t}, \V{c}, \V{p}^*)$$

\subsection{BiLSTM for scope resolution}
The scope resolution model accepts as input the token sequence $\V{t}$ and a cue vector $(c_1~\cdots~c_n)\in \{0,1\}^n$, where $c_k=0$ if the (gold or predicted) cue label of token $k$ is \textbf{NC} and $c_k=1$ otherwise. The embedding layer yields a cue embedding $\V{q} \in \{1\}^{d}$ if $c_{k}=1$ and $\V{q} \in \{0\}^{d}$ if $c_{k}=0$. For the token input, we use the same embedding matrix $E^{v\times d}$ as in the previous model.

The token and cue embeddings are passed to a BiLSTM layer with $2U$ units.\footnote{See \ref{lstm_app} for a description of a two-input LSTM cell.} An LSTM layer is well-suited for the scope resolution, since it can capture long term dependencies between a cue token and a scope token. The bidirectionality accounts for the fact that a scope token can be located to the left and the right of a cue token. The hidden state of the forward layer and backward layer are concatenated to yield a representation $\overleftrightarrow{\V{h}}_{k}=(\overrightarrow{\V{h}}_k;\overleftarrow{\V{h}}_k) \in \R^{2u}$ for token $k$. 

For each token, the output $\overleftrightarrow{\V{h}}_{k}$ of the BiLSTM layer is fed into a 4-unit dense layer with weights $W_s^{2\times 2U}$ and bias $\V{b}_s \in \R^2$. The output vector $$\V{y}_k =  W_{s} \overleftrightarrow{\V{h}}_{k} + \V{b}_s = (y_k^O, y_k^B, y_k^C, y_k^A)$$ contains to the `confidence' scores of the possible scope labels. These scores are used to obtain the final prediction label $p_k = \mathrm{softmax}(\V{y}_k)$.

\subsection{BiLSTM + CRF for scope resolution}
A BiLSTM+CRF model is also used for the scope resolution task. The model might learn that certain sequences are impossible, for example, that a \textbf{B} will never follow a \textbf{C}. Moreover, we expect that the model will yield more continuous scope predictions.

\subsection{Model training}
The objective of the models is to maximize the likelihood $\mathcal{L}(\boldsymbol{\Theta})$ of the correct predictions $\V{p}$ compared to the gold labels $\V{g}=(g_1~\cdots~g_n)$, with $\boldsymbol{\Theta}$ the set of trainable model parameters and $\V{X}$ the inputs of the model. For the BiLSTM models, this likelihood is
$$\mathcal{L}(\boldsymbol{\Theta})=\prod_{k=1}^{n} \big(p_{k}(\boldsymbol{\Theta}, \V{X})\big)^{g_t} \big(1-p_{k}(\boldsymbol{\Theta}, \V{X})\big)^{1-g_{t}},$$ 

for the BiLSTM-CRF models, this likelihood is
$$\mathcal{L}(\boldsymbol{\Theta})=\frac{e^{S(\V{X}, \V{p})}}{\sum\limits_{\V{p}^* \in \mathcal{P}} e^{S(\V{X}, \V{p}^*)}}.$$

\paragraph{Hyperparameters} The models were compiled and fitted with the Keras functional API for TensorFlow 2.3.1 in Python 3.7.6 \cite{tensorflow, python}. Based on validation results, we selected the Adam optimizer \cite{adam} with an initial learning rate 0.001 with step decay to find optimal values for $\boldsymbol{\Theta}$. Scope resolution models were trained on 30 epochs with a batch size of 32. The cue detection models were trained with early stopping, since the model showed large overfitting on 30 epochs. For the architecture hyperparameters, we selected embedding dimension $d=200$ and number of units in the LSTM-layer $U=200$. Embeddings were not updated during training, except for the cue detection baseline model. 
\subsection{Post-processing}
In Task 2, we apply a post-processing algorithm on the predictions of the BiLSTM model to obtain continuous scope predictions \cite{morante2008}. We first ensure that the cue tokens are labeled as a scope token. In case of a discontinuous negation cue, the tokens between the cue tokens are also labeled as a scope token. The algorithm locates the continuos predicion `block' containing the cue token and decides whether to connect separated blocks around it, based on their lengths and the gap length between them. 

\section{Experiments}

\subsection{Corpus} The current study made use of the Abstracts and Full papers sub corpora from the open access BioScope corpus \cite{bioscope}. Together, these sub corpora contain 14,462 sentences. For each sentence, the negation cue and its scope are annotated such that the negation cue is as small as possible, the negation scope is as wide as possible and the negation cue is always part of the scope. Resulting from this strategy, every negation cue has a scope and all scopes are continuous. 

One sentence contained two negation instances. We represented this sentence twice, such each copy corresponded to a different negation instance. This resulted in 2,094 (14.48\%) negation instances. A description of the sub corpora is provided in Table \ref{datadescription}.

\begin{table}[h]
	\centering
	\small
	\begin{threeparttable}
	\caption{Descriptive statistics of the sub corpora.}
	\label{datadescription}
		\begin{tabular}{|l | l | r|  r |}
		\hline
		\cellcolor{gray!50} & \cellcolor{gray!50}Statistic & \cellcolor{gray!50}Abstracts & \cellcolor{gray!50}Full Papers \\
		\hline
		\multirow{5}{*}{Total} 	& Documents 			& 1,273 			& 9 \\ \cline{2-4}
		& 						Sentences 			& 11,994 		& 2,469 \\ \cline{2-4}
		& 						Negation instances 	& 14.3\% 		& 15.2\% \\ \cline{2-4}
		& 						Tokens 				& 317,317 		& 69,367 \\ \cline{2-4}
		& 						OOV 		 			& 0.1\% 			& 1.4\% \\ 
		\hline
		\multirow{4}{*}{Sentence length $n$} & $n\le25$  	& 53.5\% 	& 50.6\% \\ \cline{2-4}
		& 						$25<n\le50$   		& 43.2\%			& 42.7\% \\ \cline{2-4}
		&						$50<n\le75$			& 3.0\%			& 5.6\% \\ \cline{2-4}
		&						$75<n$				& 0.3\%			& 1.1\% \\
		\hline
		\multirow{4}{*}{Scope length $S$} & $S\le10$ & 69.9\% 		& 72.0\% \\ \cline{2-4}
		& 						$10<S\le30$ 			& 24.2\% 		& 22.1\% \\ \cline{2-4}
		& 						$30<S$ 				& 58.7\%			& 58.7\% \\ \cline{2-4}
		&						Avg. $S/n$			& 0.33			& 0.30 \\
		\hline
		\multirow{5}{*}{Scope bounds} & Avg. $k_L$ 	& 16.4 			& 16.2 \\ \cline{2-4}
		&	  					Avg. $k_R$			& 23.1			& 22.8 \\ \cline{2-4}
		&	  					Avg. $k_L/n$			& 0.51			& 0.47 \\ \cline{2-4}
		&	  					Avg. $k_R/n$			& 0.76			& 0.70 \\ \cline{2-4}
		& Scope starts with cue & 85.5\%			& 78.7\% \\
		\hline
		\end{tabular}
		\begin{tablenotes}
		\small
		\item Note: OOV = Out Of Vocabulary tokens, that is, not appearing in the BioWordVec pre-trained embeddings. Avg. = average.
		\end{tablenotes}
	\end{threeparttable}
\end{table}

\paragraph{Tokenization} Biomedical text data poses additional challenges to the problem of tokenization \cite{biomedical-tokenization}. DNA sequences, chemical substances and mathematical formula's appear frequently in this domain, but are not easily captured by simple tokenizers. Examples are ``E2F-1/DP1'' and ``CD4(+)''. In the current pipeline, the standard NLTK-tokenizer was used \cite{nltk}, in accordance with the tokenizer used by the BioWordVec model. This resulted in a vocabulary of 17,800  tokens, with each token present in both sub corpora. Tokenized sentences were truncated (23 sentences) or post-padded to match a length of 100 tokens.

\subsection{Experimental set-up}

For the experiments, we apply a 70-15-15 train-validation-test split to the sub corpora. First, we train and test the cue detection models. The set of sentences with at least one predicted cue label are passed to Task 2. We use the predicted cue labels of the best model, based on the validation F1. This predicted Negation set consists of true positives and false positives: $N_{\mathrm{pred}}= tp \cup fp$. We define its complement, the predicted Assertion set, as $A_{\mathrm{pred}}=\mathrm{fn} \cup \mathrm{tn}$ and predict an empty negation scope $\V{p}\in\{\mathbf{O}\}^n$ for this set. 

\begin{figure}
	\centering
	\includegraphics[scale=0.75]{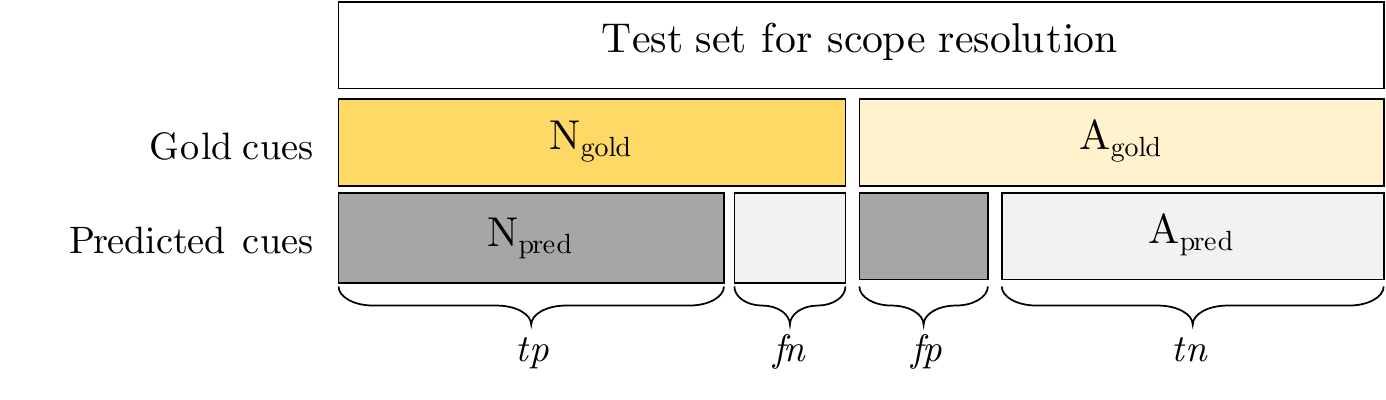}
	\caption{Visualization of negation sentences ($N$) and assertion sentences ($A$) in the test set, under different circumstances. Note: tp=true positives, fn=false negatives, fp=false postitives, tn=true negatives.}
	\label{comparison}
\end{figure}

The models in Task 2 could be tested on $N_{\mathrm{pred}}$, with predicted cue inputs. However, the model performance will be affected by the presence of false positives and absence of false negatives from Task 1 in this set. To compare this with testing on $N_{\mathrm{gold}}=tp \cup fn$ with gold cue inputs, we need to base our results on the same data. Therefore, we use $N_{\mathrm{gold}} \cup N_{\mathrm{pred}} = tp \cup fn \cup fp$ as a general test set for Task 2, see Figure \ref{comparison}. Note that $tn$ is not needed, since true negatives are not involved in the performance measures.

\section{Results and Discussion}

\subsection{Task 1 performance}
The results indicate that BiLSTM-based models can detect negation cues reasonably well in the Abstracts corpus, but perform poorly on the Full Papers corpus. The difference not surprising, since we know from previous studies that most models perform worse on the Full Papers corpus. In Table \ref{cueresults}, we report the performance of the proposed methods compared to the current state-of-the-art machine learning and neural network methods. It is clear that the models underperform on both corpora by a large margin.

The most surprising result is that none of the models perform remarkably better than the baseline model of non-trainable word embeddings. Adding a BiLSTM layer even leads to worse performace: The precision and recall measures indicate that less tokens are labeled as a cue with a BiLSTM layer, reducing the false positives, but increasing the false negatives. Appearently, the BiLSTM layer cannot capture more syntactical information needed for cue detection than already present in the embeddings. The embeddings do not benefit from a CRF layer either. It is only with a BiLSTM-CRF combination that the overall performance improves by predicting more noncue labels for tokens that are indeed not a cue token. Among the currently proposed models, we conclude that the BiLSTM+CRF model is the best for the Abstracts corpus.

In contrast, training the embeddings does lead to a better performance on the Full Papers corpus. Here, the performance measures are more conclusive. The F1 measure is halved after adding a BiLSTM layer to the embeddings, and adding a CRF leads to no predicted cue labels at all. We therefore use the trained embeddings model to obtain the cue predictions for the Full Papers corpus.

\begin{table}[h!]
	\centering
	\footnotesize
	\begin{threeparttable}
	\caption{Performance of the cue detection models.} \label{cueresults}
	\begin{tabular}{|l | c | c | c | c |}
	\hline
	\multicolumn{5}{|c|}{\cellcolor{gray!50} BioScope Abstracts} \\ \hline
	
	\cellcolor{gray!25}Method & \cellcolor{gray!25}P	& \cellcolor{gray!25}R & \cellcolor{gray!25}F1 & \cellcolor{gray!25}PECM	\\
	\hline
	Baseline & 80.59 & 87.81 & 84.05 & 76.95 \\
	\hline
	Emb. train (E) & 79.87 & 89.61 & 84.46 & 74.22 \\
	\hline
	E + BiLSTM 		& 84.87		& 82.44 	& 83.64 & 78.52	\\
	\hline
	E + CRF & 82.62 & 83.51 & 83.07 & 76.95 \\
	\hline
	E + BiLSTM + CRF  &  83.22		& 87.10  	& 85.11  &  80.86	\\
	\hline
	Metalearner \cite{morante2009} & \textbf{100}  & \textbf{98.75} & \textbf{99.37} & \textbf{98.68} \\
	\hline
	
	NegBERT \cite{negbert} & NR  & NR & 95.65 & NR \\
	\hline
	
	\multicolumn{5}{|c|}{ }  \\ \hline
	\multicolumn{5}{|c|}{\cellcolor{gray!50} BioScope Full Papers}  \\ \hline
	
	\cellcolor{gray!25}Method & \cellcolor{gray!25}P	& \cellcolor{gray!25}R & \cellcolor{gray!25}F1 & \cellcolor{gray!25}PECM	\\
	\hline
	Baseline & 64.18 & 62.32 & 63.24 & 47.46 \\
	\hline
	Emb. train (E) & 60.23 & 76.81 & 67.52 & 49.15 \\
	\hline
	E + BiLSTM 		& 58.33		& 20.28 	& 30.11 & 18.64	\\
	\hline
	E + CRF & NaN & 0 & NaN & 0 \\
	\hline
	E + BiLSTM + CRF  &  60.53		& 66.67  	& 63.45  &  45.76	\\
	\hline
	Metalearner \cite{morante2009} & \textbf{100}  & \textbf{95.72} & \textbf{96.08} & \textbf{92.15} \\
	\hline
	
	NegBERT \cite{negbert} & NR  & NR & 90.23 & NR \\
	\hline
	
	\end{tabular}
	\begin{tablenotes}
		\scriptsize
		\item Note: PECM=Percentage Exact Cue Matches.
	\end{tablenotes}
	\end{threeparttable}
\end{table}

\subsection{Task 2 performance}

Overall, it is clear that the models suffer from imperfect cue information. The F1 on the scope resolution task can decrease up to 9\% on the Abstracts corpus and 18\% on the Full Papers corpus, when moving from gold to predicted information, see Table \ref{effect}. The BiLSTM model seems to be the most robust against this effect. The transition scores of a CRF layer might make the model more receptive to cue inputs. When the model is presented a false postive cue, the transition score from an \textbf{O}-label to a \textbf{C} makes it easier to predict a false positive \textbf{C}. It is also clear why the post-processing algorithm performs worse with imperfect cue information, as it guarantees that all false positive cues will reveive a false positive scope label. This is confirmed by the sharp drop in precision (14\%) and the small drop in recall (4\%), see Table \ref{scoperesults}.

\begin{table}
\centering
\footnotesize
	\caption{F1 scores on the scope resolution task with Gold versus Predicted cue inputs.} \label{effect}
	\label{scoperesults}
	\begin{tabular}{|l|c|c|c|}
	\hline
	\multicolumn{4}{|c|}{\cellcolor{gray!50}Abstracts, Cue detection F1 = 85.11} \\ \hline
	
	\cellcolor{gray!25}Method & \cellcolor{gray!25}Gold input & \cellcolor{gray!25}Predicted input & \cellcolor{gray!25}Difference \\ \hline

	BiLSTM 		& 90.25 & 83.90 & \textbf{6.35} \\ \hline
	BiLSTM+CRF 	& 91.58 & 84.43 & 7.15 \\ \hline
	BiLSTM+post & 90.17 & 80.87 & 9.30 \\ \hline
	
	\multicolumn{4}{|c|}{\cellcolor{gray!50}Full Papers, Cue detection F1 = 67.52} \\ \hline
	
	\cellcolor{gray!25}Method & \cellcolor{gray!25}Gold input & \cellcolor{gray!25}Predicted input & \cellcolor{gray!25}Difference \\ \hline
	
	BiLSTM 		& 72.80 & 56.98 & \textbf{15.82} \\ \hline
	BiLSTM+CRF 	& 76.10 & 59.19 & 16.91 \\ \hline
	BiLSTM+post & 73.29 & 54.79 & 18.50 \\ \hline

	\end{tabular}
\end{table}

As a secondary aim, we investigated the effect of the CRF layer and the post-processing algorithm on the Percentage of Correct Scopes. In all cases, we see that the post-processing algorithm yields the highest PCS. However, this comes at the cost of a lower F1 measure at the token level when the model receives predicted cue inputs. Another disadvantage of this approach is that is not easily transferable to genres where the annotation style is different. For example, discontinuous scopes are quite common in the Conan Doyle corpus \cite{conandoyle}. 

The results indicate that the BiLSTM+CRF model often resolves more scopes completely than the BiLSTM model. This could be partly explained by the increase in continuous predictions, as earlier suggested by Fancellu et al. \cite{fancellu2017}. However, on the Full Papers corpus with predicted inputs, the CRF-based model yields a lower PCS. The precision and recall measures indicate that the BiLSTM+CRF model predicts more positive cue labels, which may result in scopes that are too wide. We also see that there remains a substantive percentage of discontinuous predictions. This may be solved by higher-order CRF layers, that is, including transitions of label $k$ to label $k+2$.

\begin{table}
\centering
\footnotesize
\begin{threeparttable}
	\caption{Performance of the scope resolution model on the Abstracts corpus.} \label{scoperesults}
	\label{scoperesults}
	\begin{tabular}{|l|l|c|c|c|c|c|}
	\hline
	\multicolumn{7}{|c|}{\cellcolor{gray!50} BioScope Abstracts}  \\ 
	
	\hline
	\cellcolor{gray!25} Cues & \cellcolor{gray!25} Method & \cellcolor{gray!25} P & \cellcolor{gray!25} R & \cellcolor{gray!25} F1 & \cellcolor{gray!25} PCS & \cellcolor{gray!25} PCP \\
	\hline
	
	\multirow{5}{*}{Gold} & BiLSTM & 89.80 & 90.70 & 90.25 & 68.34 & 87.89 \\ \cline{2-7}
	& BiLSTM+CRF & 91.07 & \textbf{92.10} & 91.58 & 70.31 & 92.19 \\ \cline{2-7}
	& BiLSTM+post & 90.43 & 89.92 & 90.17 & 72.66 & 100 \\ \cline{2-7}
	& Metalearner \cite{morante2009} & 90.68 & 90.68 & 90.67 & 73.36 & 100 \\ \cline{2-7}
	& RecurCRFs* \cite{fei2020} & \textbf{94.9} & 90.1 & 93.6 & \textbf{92.3} & - \\ \cline{2-7}
	& NegBERT \cite{negbert} & NR & NR & \textbf{95.68} & NR & NR \\
	
	\hline
	\multirow{4}{*}{Pred} & BiLSTM & \textbf{81.83} & 86.08 & 83.90 & 58.59 & 83.07 \\ \cline{2-7}
	& BiLSTM+CRF & 81.29 & \textbf{87.82} & \textbf{84.43} & 58.98 & 87.40 \\ \cline{2-7}
	& BiLSTM+post & 76.40 & 85.90 & 80.87 & 60.55 & 100 \\ \cline{2-7}
	& Metalearner \cite{morante2009} & 81.76 & 83.45 & 82.60 & \textbf{66.07} & 100 \\ \cline{2-7} 
	\hline  
	
	\multicolumn{7}{|c|}{ }  \\ \hline
	\multicolumn{7}{|c|}{\cellcolor{gray!50} BioScope Full Papers}  \\ 
	
	\hline
	\cellcolor{gray!25} Cues & \cellcolor{gray!25} Method & \cellcolor{gray!25} P & \cellcolor{gray!25} R & \cellcolor{gray!25} F1 & \cellcolor{gray!25} PCS & \cellcolor{gray!25} PCP \\
	\hline
	
	\multirow{4}{*}{Gold} & BiLSTM & 94.21 & 59.31 & 72.80 & 28.81 & 88.14 \\ \cline{2-7}
	& BiLSTM+CRF & 80.87 & 71.86 & 76.10 & 32.20 & 89.83 \\ \cline{2-7}
	& BiLSTM+post & \textbf{94.86} & 59.72 & 73.29 & 32.20 & 100 \\ \cline{2-7}
	& Metalearner \cite{morante2009} &  84.47 & \textbf{84.95} & 84.71 & \textbf{50.26} & 100 \\ \cline{2-7}
	& NegBERT \cite{negbert} & NR & NR & \textbf{87.35} & NR & NR \\  
	
	\hline
	\multirow{4}{*}{Pred} & BiLSTM & 67.69 & 49.19 & 56.98 & 18.64 & 56.92 \\ \cline{2-7}
	& BiLSTM+CRF & 57.55 & 60.93 & 59.19 & 16.95 & 63.08 \\ \cline{2-7}
	& BiLSTM+post & 49.92 & 60.73 & 54.79 & 22.03 & 100 \\ \cline{2-7}
	& Metalearner \cite{morante2009} & \textbf{72.21} & \textbf{69.72} & \textbf{70.94} & \textbf{41.00} & 100 \\ \cline{2-7} 
	\hline             
                                    
	\end{tabular}
	\begin{tablenotes}
		\scriptsize
		\item Note: PCS = Percentage Correct Scopes, PCP=Percentage Continuous scope Predictions. *These results were reported for the complete BioScope corpus.
	\end{tablenotes}
\end{threeparttable}
\end{table}

\section{Conclusion and future research}
The current study adopted a neural network-based approach to both sub tasks of negation resolving: cue detection and scope resolution. In this way, the task would be completely independent of hand-crafted features, and would more realistically demonstrate the performance on the scope detection task. The study showed that the applicability of the BiLSTM approach does not extend to cue detection: isolated word embeddings are just as effective. These embeddings could capture features that are informative for cue detection, but they need more `flexible' contextual information to distinguish negative or neutral use of a potential cue token within a given sentence. There are various architectures avaiable that could tackle this problem more effectively: Encoder-Decoder LSTMs \cite{encoderdecoder}, attention based architectures \cite{chen2019, negbert, britto2020}, hierarchical LSTMs and Embeddings from Language Models (ELMo, \cite{ELMo}).

The scope resolution performance of a BiLSTM+CRF-based method with inaccurate cue labels is hopeful. The model  still outperforms most early methods, and performs on par with some recent methods. It would be interesting to assess the robustness of other neural network-based models agains imperfect cue inputs, possibly with different levels and forms of cue accuracy. Additionally, this robustness could be integrated in the approach. For example, we could capture the prediction uncertainty of the cue inputs by feeding the probabilities instead of the labels to the scope resolution model.

We recommend researchers to adopt a two-step approach on negation resolving with neural network-based models, to [\underline{avoid} the dependency on manually created features or unrealistic cue assumptions].

\newpage 
\appendix

\section{Related Work performance} \label{relatedworkappendix}

\begin{table}[h!]
\begin{adjustwidth}{-2.5cm}{}
	\centering
	\scriptsize
	\begin{threeparttable}
	\caption{Performance of existing methods for negation cue detection.}
	\begin{tabular}{| l| l | c | c | c | c | c | c | c | c | c|}
	\hline
	\multirow{2}{*}{Approach} & \multirow{2}{*}{Method} & \multirow{2}{*}{Corpus} & \multicolumn{3}{c|}{Cue detection} & \multicolumn{5}{c|}{Scope resolution}	\\ \cline{4-11}
	& & & P & R & F1 & P & R & F1 & PCS & Cue input \\
	\hline 
	\multirow{3}{*}{RB} & Lexicon \cite{carrillo2012} & CD & 89.26 & 91.29 & 90.26 & 85.37 & 68.53 & 76.03 & 46.59 & Pred \\ \cline{2-11}
	& Lexicon \cite{ballesteros2012} & CD & 81.34 & 64.39 & 71.88 & 58.30 & 67.70 & 62.65 & 38.55 & Pred \\ \cline{2-11}
	& Formal Sem. & GMB & 88.89 & 84.85 & 86.82  & 69.20 & 82.27 & 75.17 & 40.96 & Pred \\ 
	\hline
	\multirow{12}{*}{ML} & \multirow{2}{*}{Memory-based \cite{morante2008}} & \multirow{2}{*}{BA} & \multirow{2}{*}{89.77} & \multirow{2}{*}{93.38} & \multirow{2}{*}{91.54} & 88.63 & 88.17 & 88.40 & 57.33 & Gold \\ \cline{7-11}
	& & & & & & 80.70 & 81.29 & 80.99 & 50.05 & Pred \\ \cline{2-11}
	& \multirow{6}{*}{Metalearner \cite{morante2009}} & \multirow{2}{*}{BA} & \multirow{2}{*}{100.00} & \multirow{2}{*}{98.75} & \multirow{2}{*}{99.37} & 90.68 & 90.68 & 90.67 & 73.36 & Gold \\ \cline{7-11}
	& & & & & & 83.45 & 82.60 & 82.60 & 66.07 & Pred \\ \cline{3-11}
	& & \multirow{2}{*}{BF} & \multirow{2}{*}{100.00} & \multirow{2}{*}{95.72} & \multirow{2}{*}{97.81} & 84.47 & 84.95 & 84.71 & 50.26 & Gold \\ \cline{7-11}
	& & & & & & 72.21 & 69.72 & 70.94 & 41.00 & Pred \\ \cline{3-11}
	& & \multirow{2}{*}{BC} & \multirow{2}{*}{100.00} & \multirow{2}{*}{98.09} & \multirow{2}{*}{99.03} & 91.65 & 92.50 & 92.07 & 87.27 & Gold \\ \cline{7-11}
	& & & & & & 86.38 & 82.14 & 84.20 & 70.75 & Pred \\ 
	\cline{2-11}
	& Lexicon+Support Vector Machine \cite{gyawali2012} & CD & 85.93 & 85.61 & 85.77 & 85.37 & 68.86 & 76.23 & 53.01 & Pred \\
	\cline{2-11}
	& Support Vector Machine \cite{read2012} & CD & 91.42 & 92.80 & 92.10 & 81.99 & 88.81 & 85.26 & 61.45 & Pred \\ \cline{2-11}
	& \multirow{2}{*}{MRS Crawler \cite{packard2014}} & \multirow{2}{*}{CD} & \multirow{2}{*}{-} & \multirow{2}{*}{-} & \multirow{2}{*}{-} & 86.4 & 86.8 & 86.6 & 70.2 & Gold \\ \cline{7-11}
	& & & & & & 80.0 & 84.9 & 82.4 & 67.9 & Pred* \\ \hline 
	\multirow{2}{*}{CRF} & CRF \cite{jbara2012} & CD & 94.31 & 87.88 & 90.98 & 84.85 & 80.66 & 82.70 & 50.60 & Pred \\ \cline{2-11}
	& CRF \cite{white2012} & CD & 88.04 & 92.05 & 90.00 & 83.26 & 83.77 & 83.51 & NR & Pred \\ \hline 
	\multirow{12}{*}{NN} & \multirow{3}{*}{CNN \cite{qian2016}} & BA & - & - & - & 89.49 & 90.54 & 89.91 & 77.14 & Gold \\ \cline{4-11}
	& & BF & - & - & - & 82.08 & 84.90 & 83.46 & 53.99 & Gold \\ \cline{4-11}
	& & BC & - & - & - & 91.97 & 97.03 & 94.43 & 87.82 & Gold \\ \cline{2-11}
	& BiLSTM \cite{fancellu2016} & CD & - & - & - & 92.62 & 85.13 & 88.72 & 63.87 & Gold \\ \cline{2-11}
	& \multirow{2}{*}{BiLSTM+CRF \cite{fancellu2017}} & BA & - & - & - & NR & NR & 92.11 & 81.38 & Gold \\ \cline{4-11}
	& & BF & - & - & - & NR & NR & 77.73 & 54.54 & Gold \\ \cline{2-11}
	& LSTM Encoder-Decoder \cite{gautam2018} & DT  & 100 & 99.8 & 99.9 & 85.0 & 82.6 & 83.7 & NR & None \\ \cline{2-11}
	& BiLSTM \cite{taylor2018} & BA & NR & NR & NR & 88.72 & 89.02 & 88.85 & NR & None \\ \cline{2-11}
	& RecurCRF \cite{fei2020} & B & - & - & - & 94.9 & 90.1 & 93.6 & 92.3 & Gold \\ \cline{2-11}
	& \multirow{3}{*}{NegBERT \cite{negbert}} & CD & NR & NR & 92.94 & NR & NR & 92.36 & NR & Gold \\ \cline{4-11}
	& & BA & NR & NR & 95.65 & NR & NR & 95.68 & NR & Gold \\ \cline{4-11}
	& & BF & NR & NR & 90.23 & NR & NR & 87.35 & NR & Gold \\ \hline
	\end{tabular}
	\begin{tablenotes}
		\scriptsize
		\item Note: CD = Conan Doyle \cite{conandoyle}, GMB = Groningen Meaning Bank \cite{groningendata}, B(A,F,C) = BioScope (Abstracts, Full Papers, Clinical) \cite{bioscope}, DT = DeepTutor Negation \cite{deeptutor}. *Predictions from SVM \cite{read2012}. NR = Not Reported. A dash indicates that the cue detection task was not performed.
	\end{tablenotes}
	\end{threeparttable}
\end{adjustwidth}
\end{table}

\newpage

\section{Motivation of the scope labeling scheme} \label{motivation}
The scope labeling scheme was motivated by the transition scores in a CRF model. Let $T^{5\times5}$ be a matrix such that $T_{i,j}$ represents a score associated with predicting label $i$ for $t_k$ and label $j$ for $t_{k+1}$. Based on the structure of a scope within a sentence, we could expect the following kind of structure within $T$, where $-2=$ impossible, $-1=$ unlikely, $1=$ likely, $2=$ very likely:

$$T=\begin{pmatrix}
 			& \mathbf{O} & \mathbf{B} & \mathbf{C} & \mathbf{A} \\
\mathbf{O}	& 1			 & 1 	     & 1  		  & -2  \\
\mathbf{B}	& -2			 & 1 	     & 1  		  & -2  \\
\mathbf{C}	& -1  		 & -2 	     & -1  		  & 2  \\
\mathbf{A}	& 1   		 & -2 	     & -2  		  & 1  \\
\end{pmatrix}
$$

\newpage

\section{A LSTM-cell with two inputs $\V{e}_k, \V{q}_k \in \R^d$} \label{lstm_app}
\begin{figure}[h!]
\centering
	\includegraphics[scale=0.6]{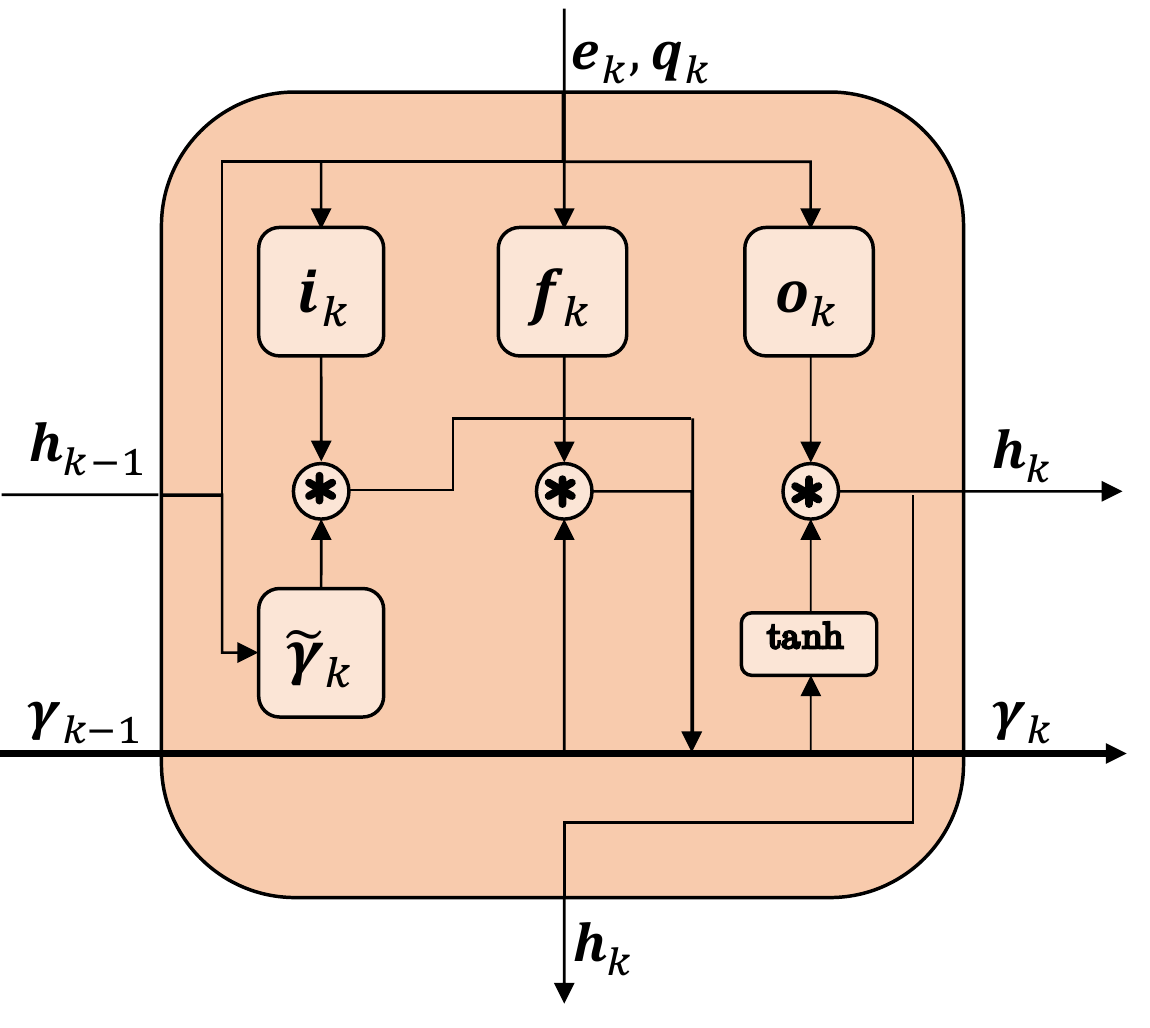}
	\caption{Visualization of the LSTM cell architecture.}
	\label{cell}
\end{figure}

An LSTM cell with two inputs is given by the following equations:

$$
\begin{aligned}
\mathbf{i}_k &= \sigma\big(W_e^{(i)}\V{e}_k + W_q^{(i)}\V{q}_k + W_h^{(i)}\V{h}_{k-1} +\V{b}^{(i)} \big), \\ 
\V{f}_k &= \sigma\big(W_e^{(f)}\V{e}_k + W_q^{(f)}\V{q}_k
+ W_h^{(f)}\V{h}_{k-1} +\V{b}^{(f)} \big), \\
\V{o}_k &= \sigma\big(W_e^{(o)}\V{e}_k + W_q^{(o)}\V{q}_k
+ W_h^{(o)}\V{h}_{k-1} +\V{b}^{(o)} \big), \\
\tilde{\boldsymbol{\gamma}}_k&=\mathrm{tanh}\big(W_e^{(\tilde{\boldsymbol{\gamma}})}\V{e}_k +
W_q^{(\tilde{\boldsymbol{\gamma}})}\V{q}_k + W_h^{(\tilde{\boldsymbol{\gamma}})}\V{h}_{k-1} + \V{b}^{(\tilde{\boldsymbol{\gamma}})}\big), \\
\boldsymbol{\gamma}_k&=\mathbf{f}_k * \boldsymbol{\gamma}_{k-1} + \V{i}_k * \tilde{\boldsymbol{\gamma}}_{k}, \\
\V{h}_k&=\mathbf{o}_k * \mathrm{tanh}(\boldsymbol{\gamma}_k),
\end{aligned}
$$

where $W_e^{U \times d}$ and $W_q^{U \times d}$ denote weight matrices for the token and cue embeddings respectively, $W_{h}^{U\times U}$ denotes the recurrent weight matrix, $\V{b} \in \R^{u}$ is a bias vector, $\odot$ denotes the Hadamard product, $\sigma$ denotes the sigmoid function $\R \rightarrow (0,1)$ given by $x \mapsto 1/(1+e^{-x})$ and tanh denotes the hyperbolic tangent function $\R \rightarrow (-1,1)$ given by $x \mapsto (e^x-e^{-x})/(e^x+e^{-x})$.

\newpage

\end{document}